\documentclass[conference]{IEEEtran}
\IEEEoverridecommandlockouts
\usepackage{cite}
\usepackage{amsmath,amssymb,amsfonts}
\usepackage[hyphens,spaces,obeyspaces]{url}
\usepackage{algorithmic}
\usepackage{graphicx}
\usepackage{textcomp}
\usepackage{xcolor}
\usepackage{booktabs}
\usepackage{color}
\usepackage{algorithm,algorithmic}
\usepackage{subcaption}
\usepackage{xspace}
\newcommand{\ie}{{i.e.}}

\newcommand{\bt}{\textsc{BT}\xspace}
\newcommand{\ftonly}{\textsc{FT}\xspace}
\newcommand{\ft}[1]{FT-Partial{#1}\xspace}
\newcommand{\itonly}{\textsc{IT}\xspace}
\newcommand{\itadd}{\textsc{IT-Add}\xspace}
\newcommand{\itpad}{\textsc{IT-Pad}\xspace}
\newcommand{\cifards}{CIFAR100\xspace}
\newcommand{\fiveds}{FIVEDS\xspace}
\newcommand{\cifartask}{CIFAR100/T10\xspace}
\newcommand{\resisctask}{RESISC45/T9\xspace}
\newcommand{\fivedstask}{FIVEDS/T5\xspace}
\newcommand{\domnettask}{DOMAINNET/T4\xspace}
\newcommand{\singlesource}{Single Source Data\xspace}
\newcommand{\multisource}{Multiple Source Data\xspace}

\usepackage{array}

\usepackage{lipsum,graphicx,multicol}

\makeatletter
\newcommand{\linebreakand}{%
  \end{@IEEEauthorhalign}
  \hfill\mbox{}\par
  \mbox{}\hfill\begin{@IEEEauthorhalign}
}
\makeatother

\newcolumntype{H}{>{\setbox0=\hbox\bgroup}c<{\egroup}@{}}

\def\BibTeX{{\rm B\kern-.05em{\sc i\kern-.025em b}\kern-.08em
    T\kern-.1667em\lower.7ex\hbox{E}\kern-.125emX}}
\begin{document}

\title{Continual Learning with Pretrained Backbones\\ by Tuning in the Input Space
\thanks{This work was partly supported by the PRIN 2017 project RexLearn, funded by the Italian Ministry of Education, University and Research (grant no. 2017TWNMH2). This work was also partially supported by TAILOR and by HumanE-AI-Net, projects funded by EU Horizon 2020 research and innovation programme under GA No 952215 and No 952026, respectively.  

Accepted for publication at the IEEE International Joint Conference on Neural Networks (IJCNN) 2023 (DOI: TBA).
\copyright 2023 IEEE. Personal use of this material is permitted. Permission from IEEE must be obtained for all other uses, in any current or future media, including reprinting/republishing this material for advertising or promotional purposes, creating new collective works, for resale or redistribution to servers or lists, or reuse of any copyrighted component of this work in other works.}}
\author{

\IEEEauthorblockN{Simone Marullo\IEEEauthorrefmark{1}\IEEEauthorrefmark{2}}
\IEEEauthorblockA{\IEEEauthorrefmark{1}\textit{DINFO} \\
\textit{University of Florence}\\
Florence, Italy\\
\footnotesize{\texttt{simone.marullo@unifi.it}}}
\and
\IEEEauthorblockN{Matteo Tiezzi\IEEEauthorrefmark{2}}
\IEEEauthorblockA{\IEEEauthorrefmark{2}\textit{DIISM} \\
\textit{University of Siena}\\
Siena, Italy\\
\footnotesize{\texttt{matteo.tiezzi@unisi.it}}}
\and
\IEEEauthorblockN{Marco Gori\IEEEauthorrefmark{2}\IEEEauthorrefmark{3}}
\IEEEauthorblockA{\IEEEauthorrefmark{3}\textit{MAASAI} \\
\textit{Universit\`{e} C\^{o}te d'Azur}\\
Nice, France\\
\footnotesize{\texttt{marco.gori@unisi.it}}
}
\linebreakand
\IEEEauthorblockN{Stefano Melacci\IEEEauthorrefmark{2}}
\IEEEauthorblockA{\IEEEauthorrefmark{2}\textit{DIISM} \\
\textit{University of Siena}\\
Siena, Italy\\
\footnotesize{\texttt{mela@diism.unisi.it}}}
\and
\IEEEauthorblockN{Tinne Tuytelaars\IEEEauthorrefmark{4}}
\IEEEauthorblockA{\IEEEauthorrefmark{4}\textit{ESAT} \\
\textit{KU Leuven}\\
Leuven, Belgium\\
\footnotesize{\texttt{tinne.tuytelaars@esat.kuleuven.be}}}
}

\maketitle

\begin{abstract}
The intrinsic difficulty in adapting deep learning models to non-stationary environments limits the applicability of neural networks to real-world tasks. This issue is critical in practical supervised learning settings, such as the ones in which a pre-trained model computes projections toward a latent space where different task predictors are sequentially learned over time. As a matter of fact, incrementally fine-tuning the whole model to better adapt to new tasks usually results in catastrophic forgetting, with decreasing performance over the past experiences and losing valuable knowledge from the pre-training stage. In this paper, we propose a novel strategy to make the fine-tuning procedure more effective, by avoiding to update the pre-trained part of the network and learning not only the usual classification head, but also a set of newly-introduced learnable parameters that are responsible for transforming the input data. This process allows the network to effectively leverage the pre-training knowledge and find a good trade-off between plasticity and stability with modest computational efforts, thus especially suitable for on-the-edge settings. 
Our experiments on four image classification problems in a continual learning setting confirm the quality of the proposed approach when compared to several fine-tuning procedures and to popular continual learning methods.
\end{abstract}


\section{Introduction}
\label{sec:intro}

The outstanding performance achieved by Machine Learning solutions in a vast variety of fields \cite{gpt3} and well-defined tasks \cite{objectrec} is usually restricted to a very specific setting,
where it's assumed that all training data is available from the start and sampled from a static distribution in an independent manner (i.i.d.). This scenario does not contemplate the case in which neural models are progressively adapted to novel data that is sampled over time from non-stationary distributions. 
Recently, 
the limitations implied by the i.i.d. assumption have gained wider attention and novel models that are designed to learn over time started to emerge \cite{delange2021clsurvey}. Such models are loosely inspired by human cognition, which typically works in an incremental fashion with new notions being learnt in a fruitful relation with previously acquired knowledge, so that both old and new skills are often refined through their interaction \cite{human}. This is in stark contrast with the default behavior of neural networks, where a huge performance drop on old data is typically noticed when adapting weights on novel and never-seen-before datasets presented over time.
This issue, generally referred to as \emph{catastrophic forgetting}\cite{catastrophic}, has been known since the early connectionist movement \cite{human} and is still far from being solved in a general and satisfying way. To address these issues, \emph{Continual Learning} (CL) develops methods suitable for problems in which training data are presented
over time and potentially in a lifelong
manner.

Intelligent edge devices, including sensors, actuators, robotic platforms, etc., with modest computational resources, are becoming ubiquitous and there is a growing need of Machine Learning-driven processing capabilities for perception, understanding and personalization in the Internet of Things. Indeed, such devices are capable of acquiring continuous data streams \cite{mledge}, that could be processed with CL-based solutions running on the edge devices themselves.
Several challenges must be faced when deploying CL methods to edge devices. First of all, typical state-of-the-art neural architectures (e.g.,  Transformers \cite{NIPS2017_3f5ee243}) with many encoding layers might be of limited applicability due to scarcity of memory and computational capabilities. While offloading to the cloud might be a simple workaround, it raises privacy concerns and complexity issues. 
Moreover, most of the state-of-the-art CL methods involve rehearsal procedures \cite{delange2021clsurvey}, \ie, re-visiting some exemplars of past concepts in order to ``refresh'' the knowledge of the model. This opens to new issues concerning storage capacity and, again, privacy \cite{pellegrini}, given that it introduces the need of long-term storage of training data. Methods based on latent replay (rather than input replay) achieve better compression and obfuscate potentially private data, but storage issues are not solved and privacy preservation may be totally unreliable \cite{invertingfeat}.
Another issue with the current CL research concerns the limited scale of the tackled tasks, 
with only limited efforts spent in evaluating the performance of CL methods in realistic applications. 

In the context of edge devices with limited computational capabilities, it seems reasonable to start from networks pre-trained on large data collections, since they are powerful tools to compute informed latent representations and they allow to reduce training efforts.
However, their embedded knowledge might be significantly lost when progressively fine-tuning the network to novel domains in a CL fashion (especially when rehearsal is not performed \cite{foundation}).
An alternative to fine-tuning has recently become popular in the Machine Learning community, \ie, learning parameters that affect the model input with the goal of conditioning the network rather than changing its internal weights.
These Prompt Tuning models \cite{li2021prefixtuning,lester-etal-2021-power,vpt} were conceived to foster the exploitation of very-large-scale networks, in order to effectively leverage their capabilities in specific downstream tasks. However, most of the work in this line focuses on large Transformer models and the connection with CL is not deeply investigated \cite{l2p}.
Motivated by these considerations, in this paper we propose and investigate the appropriateness of different tuning options when facing a CL problem based on a pre-trained network, frequently referred to as {\it backbone}, focusing on methods that require lower computational efforts and no-rehearsal, well suited for edge devices. In particular, we propose \textit{Input Tuning} (\textsc{IT}), an alternative form of Prompt Tuning, in order to efficiently adapt the model to new data and achieve a good trade-off between plasticity and stability. We provide an experimental analysis conducted on different datasets available in the related literature, comparing several fine-tuning procedures and well-known continual learning methods.
The contributions of this paper are the following ones: (1) we propose the adoption of Input Tuning procedures to better leverage pre-trained backbones in CL;
(2) we experimentally evaluate the impact of fine-tuning on common CL benchmarks, providing reference results; (3) we investigate and show the benefits of Input Tuning in a continual setting, 
with edge-friendly neural architectures, both for small and large domain shifts.

This paper is organized as follows. Related work is presented in Section~\ref{sec:related}, while Input Tuning is described in Section~\ref{sec:proposal}. 
Experiments are in Section~\ref{sec:experiments} and Section~\ref{sec:indepth}, while conclusions and suggestions for future work are drawn in Section~\ref{sec:conclusions}.

\section{Related works}\label{sec:related}

The wide availability of pre-trained models offers several opportunities to transfer their knowledge to specific downstream tasks. The simplest approach consists in fine-tuning the models on the novel task data. This is typically demanding in terms of resources, especially in the case of large-scale models, due to the memory occupation (gradients as well as activations) and the operations in the weight update routine.
 The generic notion of pre-training has been shown to implicitly make some CL problems easier \cite{mehta2021empirical}, following the intuition that it is more likely to end up in more informed solutions compared to randomly-initialized models, due to the skills learned in the pre-training stage. Of course, an inappropriate fine-tuning procedure might yield a model too strongly focused on the novel task, losing the advantage from the previously learned knowledge.
However, while the limits of transferring pre-trained models to downstream tasks were recently studied \cite{abnar2022exploring}, when it comes to specifically studying or evaluating the concrete impact of adapting pre-trained models to a CL context, the scientific literature is relatively scarce \cite{foundation}. 
Ramasesh et al. \cite{ramasesh2022effect} have pointed out that resistance to forgetting consistently scales with the network size when employing pre-trained models. Still, several questions remain unanswered, especially concerning smaller-scale models in computationally-restricted environments, as the ones we study in this paper.
Another line of work \cite{dapeng, cossu} replaces the pre-training step with a continual procedure, especially in the self-supervised setting. While this approach paves the way for the exploitation of decentralized and streaming data in a variety of contexts, it is not focused on the practical scenario in which downstream tasks are learnt in an incremental fashion. 

Another research topic related to this paper consists of Prompt Tuning models \cite{li2021prefixtuning, lester-etal-2021-power, inputtuning}, that were originally conceived in Natural Language Processing (NLP), and that open a new perspective on how to adapt a pre-trained model to a novel environment, related to the one of the pre-training stage. 
Prompt Tuning models learn parameters that are actually part of the input stage, in order to discover the proper way to condition the model, rather than changing the values of its internal weights or learning additional internal parameters (as in the case of Adapters \cite{pfeiffer2020AdapterHub}). Such technique has been mostly applied to Transformer models, both for language tasks and vision tasks.  In this paper we focus on a specific type of tuning that is not restricted to Transformer models, and that we study in the the context of CL.

Tweaking the input space to alter the behavior of neural networks has already been investigated for a variety of purposes. With this strategy, researchers successfully managed to re-program a trained model to perform on a very different task \cite{elsayed2018adversarial}, while others \cite{marullo2021friendly,Marullo_Tiezzi_Gori_Melacci_2022} developed a curriculum-learning inspired methodology to effectively learn from a continuous set of learning problems with gradually increasing complexity.
Recently, this idea has been studied in the context of transfer learning for image classification \cite{vpt}, and it has been shown that learning parameters to transform  the input data is a competitive approach to deal with large pre-trained models, especially Transformer-based ones.
Researchers are starting to apply this intuition to the CL perspective \cite{l2p}, but they are mostly focused on rather large ViT \cite{dosovitskiy2021an} models. Moreover, they heavily rely on the capability of the considered model to extract global semantically-rich representations to guide the input transformation, while they lack explicit comparison with the non-Transformer case. In this paper we specifically study how altering the input data by means of newly introduced learnable parameters can yield efficient and computationally affordable CL, starting from a pre-trained backbone.

\begin{figure}[!ht]
\centering
 \includegraphics[width=1.1\columnwidth]{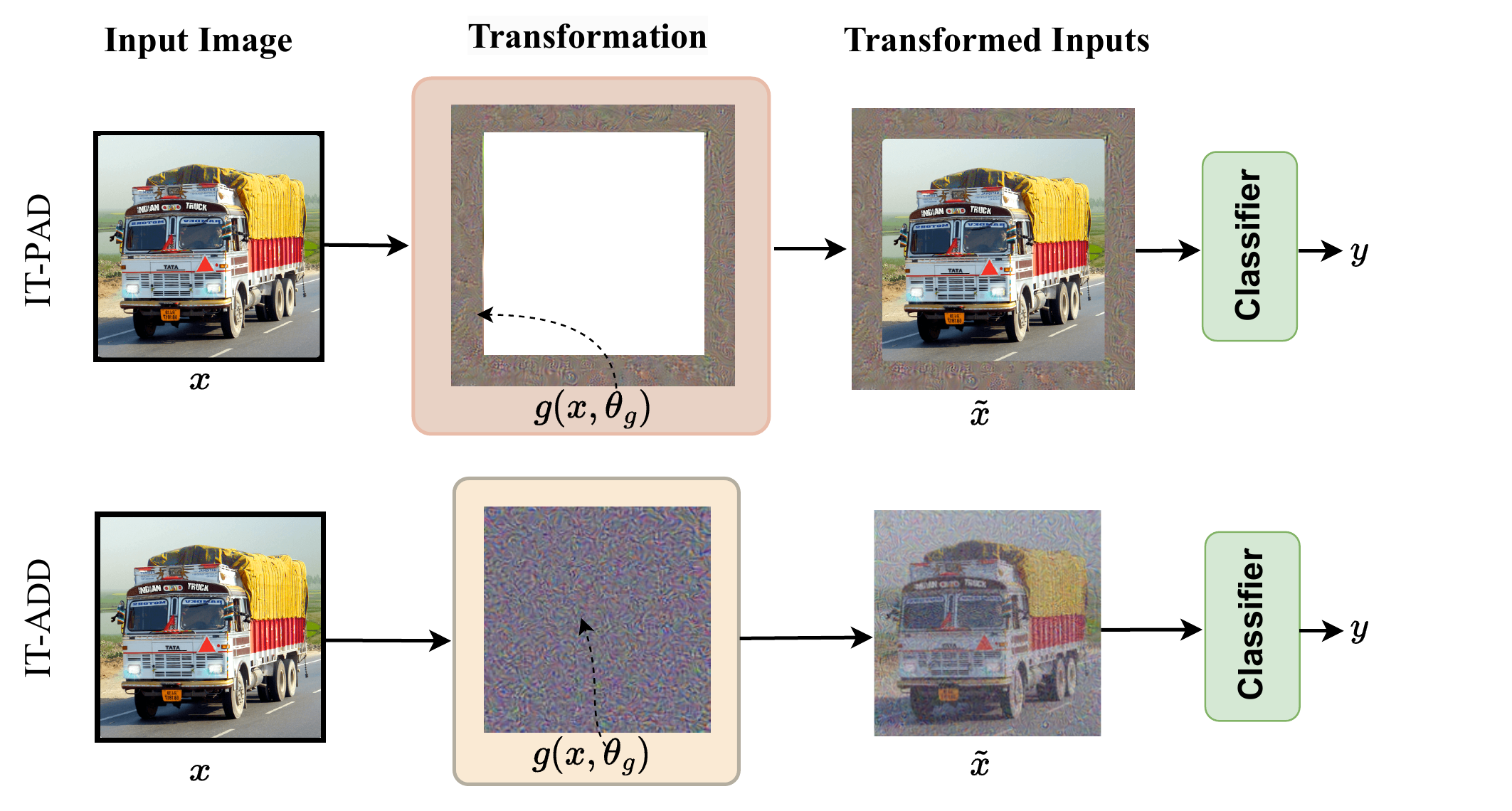}
\centering
\caption{Sketch of the proposed Input Tuning procedure. Two  variants (\itpad, \itadd) of the transformation function $g(\cdot,\cdot)$, showing examples of the newly introduced learnable parameters $\theta_g$ and of the resulting transformed inputs $\tilde{x}$. }
\label{fig:diagram}
\end{figure}

\section{Continually Tuning a Pre-trained model}\label{sec:proposal}
We consider a specific setting in which learning is performed by an edge device with limited computational resources, shipped with a neural network that was pre-trained on an initial task (usually large-scale data).
Without any loss of generality, we focus on image classification problems, so that the input of the network is a $w \times h$ RGB image.
As usual, the network can be considered to be composed by a feature extractor $m$ that extracts higher-level features, and a classification head $c$ that returns the predicted confidence $y$ on a set of classes,
$$
y = c(m(x, \theta_m), \theta_{c}),
$$
where $\theta_m$ and $\theta_{c}$ are the parameters  (weights and biases) of the feature extractor and the classification head, respectively, and $y$ is a vector with a number of components equal to the number of classes. For simplicity, we consider the classification head to be composed only of the last linear-projection and non-linearity. 
The network knowledge can be transferred to other somewhat related tasks by removing $c$ and replacing it with a task-specific head $\hat{c}$, with its own new $\theta_{\hat{c}}$. The original $m$ acts as a backbone, and $\theta_m$ can be fine-tuned on the new task, together with learning from scratch the new $\theta_{\hat{c}}$. We indicate with \ftonly such a fine-tuning approach, that can be possibly restricted to $\theta_{\hat{c}}$ and a subset of $\theta_m$. We refer to that setting as \ft{}. We further distinguish these cases from a more lightweight option called Bias Tuning (\bt), which is known to target a particularly small and expressive subset of model parameters \cite{tinytl}, i.e., the biases of the neurons of the whole network, together with the usual $\theta_{\hat{c}}$.
We assume that the net is simple enough to perform fast inference with the considered hardware, \ie, in an amount of time that is acceptable for the target scenario. 

We focus on the popular CL setting in which learning consists of a certain number ($T \geq 1$) of separated training sessions, $\mathcal{S}_j$, $j=1,\ldots,T$, where $T$ could be potentially infinite (lifelong learning). Each session comes with data $\mathcal{D}_j$, and there is no overlap between data batches of different sessions, $\mathcal{D}_j \cap \mathcal{D}_h = \emptyset$, $\forall (j,h)$. 
In each $\mathcal{S}_j$ the selected learner is presented with a task and after the end of the session the task data are no more available for further training. 
Before starting the CL process, we plug a novel classification head $\hat{c}$ on top of the pre-trained $m$, using a large number of output neurons (at least equal to the expected total number of classes at the end of the whole learning process), and randomly initializing its parameters. Of course, we assume that the pre-training dataset is sufficiently generic to be helpful for tackling a variety of downstream learning problems. 
The final goal is to effectively tune the model in a progressive manner, using data coming from the stream of tasks, without storing past information. We indicate with $\theta_{*}^{(j)}$ the values of the parameters after session $\mathcal{S}_j$, where $*$ is a placeholder for all the different parameters mentioned in this paper. 
We expect the overall model at the end of the sequence of tasks,
$$
y = \hat{c}\left(m\left(x,\theta_m^{(T)}\right), \theta_{\hat{c}}^{(T)}\right),
$$
 to have high average accuracy on all the tasks, keeping the average forgetting of knowledge at minimum. 

We will mostly focus on the class-incremental setting with some insights on the domain-incremental one, in both cases in a fully supervised scenario. 
In the {\it class-incremental} setting we are given data partitioned into a large number of classes and we simulate CL by limiting each session $\mathcal{S}_j$ to a specific subset of them (disjoint subsets). 
For each $\mathcal{S}_j$ the classifier outputs confidence scores $y$ limited to the session-related classes, thus exploiting a portion of the head $\hat{c}$, referred to as $\hat{c}_j$.\footnote{In case of softmax activation on the output units, our notation $y$ refers to the non-normalized logits (and not to the softmax output).} Gradients are only computed for the parameters that involve the output neurons in $\hat{c}_j$, whose values are indicated with $\theta_{\hat{c}_j}$.
This is an instance of the so-called ``label trick'' \cite{zeno}, a rather simple technique that has a very effective role in preventing interference and, as such, it is considered as a stable part of all the models.
In the {\it domain-incremental case} 
 the learner is presented with new data labelled over the same set of classes through the sessions, but originating from a different data population, thus the label trick does not apply. In fact, for each $\mathcal{S}_j$ the classifier outputs the full set of confidence scores using the whole $\hat{c}$, and gradients are always computed with respect to all the $\theta_{\hat{c}}$. 
In all the CL experiments of this paper we always assume that the task identity is {\it not known} at test time, since it is more challenging and more realistic. 
It is important to remark the difference of what we have described so far from the more common transfer learning process in which all the data ($\cup_{j=1}^{T}\mathcal{D}_j$) is simultaneously available to tune the model, that we refer to as Joint Learning (JL) setting, since all the new tasks are jointly as 
 processed.
As a matter of fact, learning in a CL setting is significantly more challenging than in JL, given that we expect the model to keep a reasonable performance on past tasks while learning the new ones in a sequential manner, without storing past data.

\subsection{Proposed Approach}
In this paper, we propose to consider a generic Input Tuning (\itonly) in which the original input image $x$ is transformed into $\tilde{x}$ by some function $g$ before feeding it to the network  (Fig.~\ref{fig:diagram}),
\begin{eqnarray*}
    \tilde{x} &=& g(x, \theta_g)\\
    y &=& \hat{c}(m(\tilde{x},\theta_m), \theta_{\hat{c}}),
\end{eqnarray*}
where $\theta_g$ are {\it newly introduced} learnable parameters involved in the transformation function only. The purpose of the optimization carried out during the CL process is to jointly train the parameters $\theta_{\hat{c}}$ of the classifier $\hat{c}$ together with the novel input tuning parameters $\theta_g$. 
The first Input Tuning approach we consider, which we refer to as \itpad, consists in framing the input image with a border (referred to as \emph{Frame}) of learnable ``pixels'', described by $\theta_g$. Another simple alternative, henceforth denoted as \itadd, consists in transforming the input by adding to $x$ a learnable tensor $\theta_g$ (termed as \emph{Perturbation}) of the same shape as the input image, shared by all the possible inputs. Fig. \ref{fig:diagram} shows a visual sketch of these transformations, including an example taken from the experiments that will be thoroughly discussed in Sec. \ref{sec:experiments}.

In order to provide a glimpse on the outcome of \itonly, we report in Tab. \ref{tab:cifarjoint} the change in accuracy we get when comparing a non-tuned baseline model (i.e., a model in which only $\theta_{\hat{c}}$ are trained while the whole backbone is kept fixed) with the previously introduced fine-tuning or Input Tuning procedures, both when performing transfer learning using all the data (JL, third column), and when performing sequential CL (fourth column), which is the focus of this work and will be detailed in the following. 
It is evident that what works very well in the JL setting is not necessarily well-suited for the CL case, confirming the importance of investigating alternative ways of exploiting pre-trained backbones in CL. Interestingly, the \itonly instances (\itpad and \itadd) are the ones that better perform in CL, even if they learn a relatively small set of parameters. 
\begin{table}[h]
\begin{center}
\begin{tabular}{l|l|Hl|Hl}
\toprule
\textbf{Tuning} & \textbf{Learnt Parameters} & Acc. JL & \textbf{Joint} & Acc. CL & \textbf{Continual} \\ \midrule
None & $\theta_{\hat{c}}$ & 66.12 & 66.12 & 44.49 & 44.49\\
\midrule
\bt & $\theta_{\hat{c}}$ and Biases of $m$ ($\sim$ 1k) & 70.56 & +4.44 & 41.72 & -2.77\\
\ft1 & $\theta_{\hat{c}}$ and $\theta'_{m}$ (4.7M) & 67.81 & +1.69 & 14.02 & -30.47\\
\ft2 & $\theta_{\hat{c}}$ and $\theta''_{m}$ (8.4M) & 63.96 & -2.16 & 12.29 & -32.20\\
\itpad & $\theta_{\hat{c}}$ and {\it Frame} (0.1M) & 67.98 & +1.86 & \textbf{51.58} & \textbf{+7.09}\\
\itadd & $\theta_{\hat{c}}$ and {\it Perturbation} (0.15M) & 65.56 & -0.56 & 44.81 & {+0.32}\\
\bottomrule
\end{tabular}
\caption{Preview of the impact of different tuning approaches on \cifards in the Joint (accuracy) and Continual (average task accuracy measured at the end of the learning sequence) Learning settings. The first row reports the absolute results of the baseline.
The {\it differences} with respect to them are reported in the other rows. While Bias Tuning (\bt) is very effective when all the examples are simultaneously available for training, Input Tuning is a competitive approach in the CL setting - where partial fine-tuning (\ft) badly fails. $\theta'_m$ and $\theta''_m$ are two different subsets of the backbone parameters -- see Sec. \ref{sec:experiments} for more details.
}\label{tab:cifarjoint}
\end{center}
\end{table}

\begin{figure*}[h]
\centering
\begin{subfigure}{0.48\textwidth}
\centering
\vskip -0.3cm
\includegraphics[width=0.6\textwidth]
{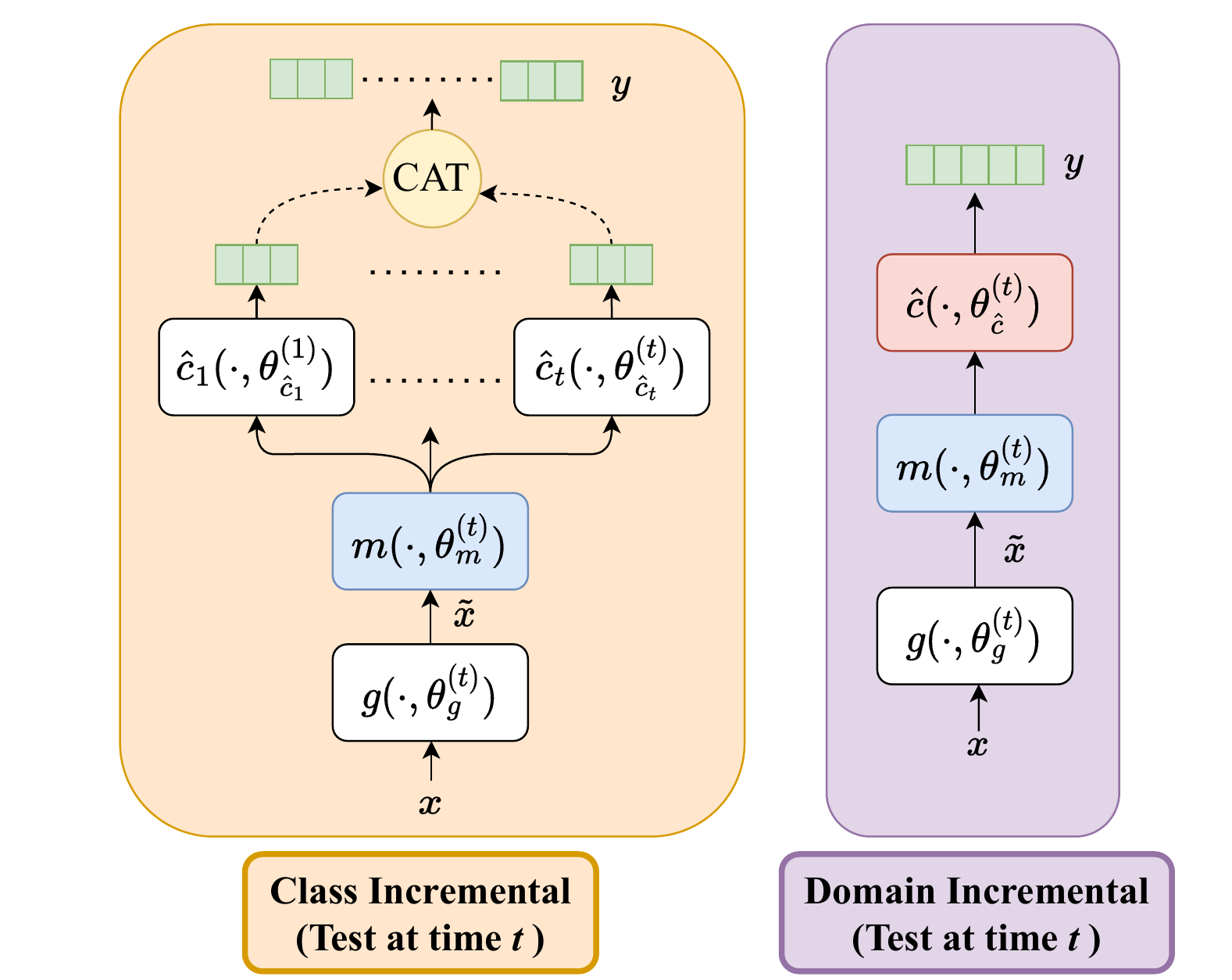}
\caption{} \label{fig:2a}
\end{subfigure}
\begin{subfigure}{0.48\textwidth}
\centering
\includegraphics[width=0.8\textwidth]{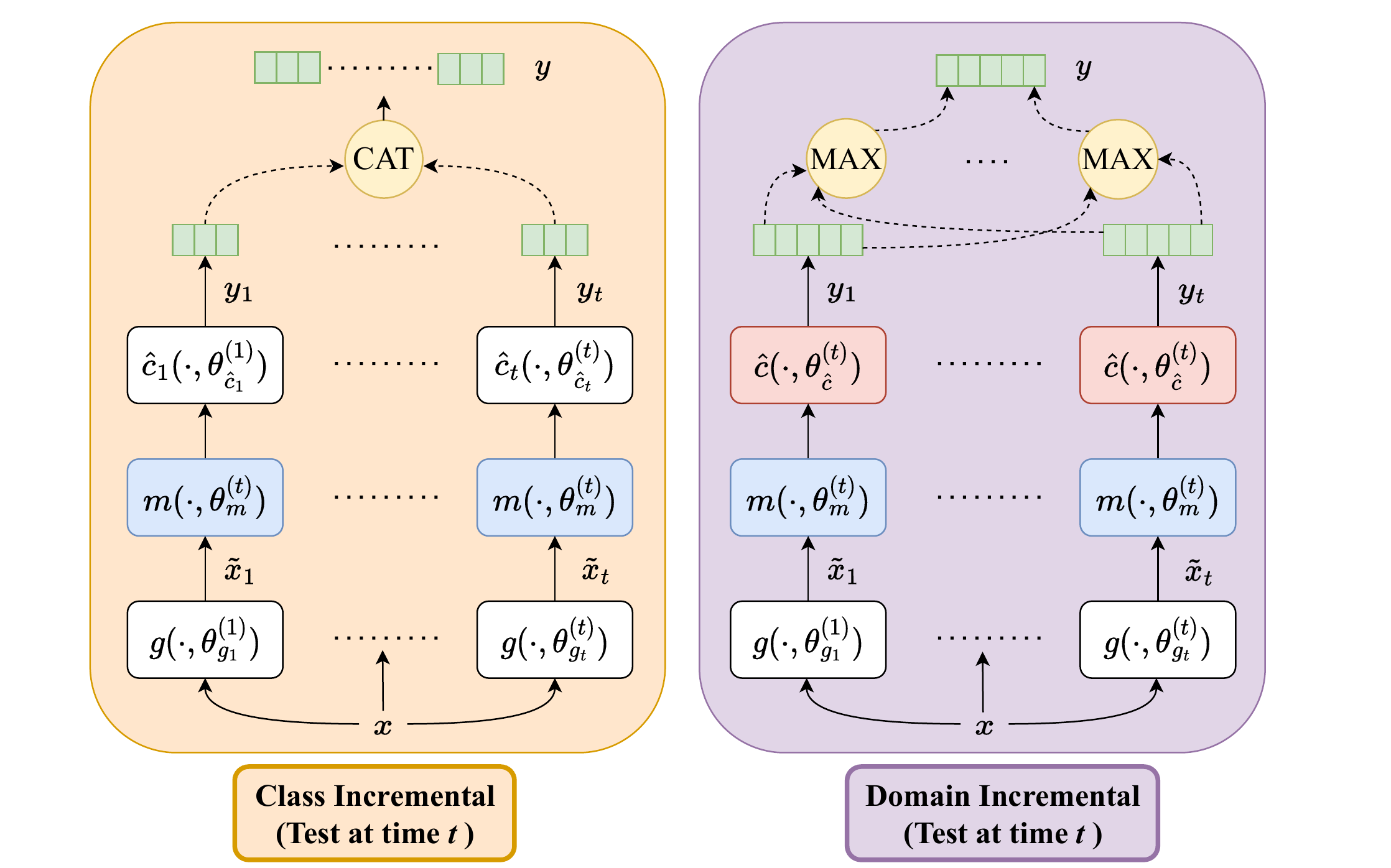}
\caption{} \label{fig:2b}
\end{subfigure}
\caption{Sketch of Input Tuning computational pipeline for the {\it class-incremental} and {\it domain-incremental} cases, going from input $x$ to class-confidence scores $y$ (logits in the case of softmax) at inference time: (a) standard approach, (b) the \emph{parallel classifier} variant. }
\label{fig:diagram2}
\end{figure*}

Whenever a continual learning problem spans clearly heterogeneous distributions (for example in the case of data collected from multiple sources), sharing the exact same transformation $g(\cdot, \theta_g)$ for all the tasks (referred to as \emph{standard approach}, depicted in Fig. \ref{fig:2a}) may not be optimal.  For this reason, we propose to learn a different input transformation in different training sessions, by learning independent $\theta_g$'s, one for each task of the session. We will indicate with $\theta_{g_j}$ the transformation parameters learnt in session $\mathcal{S}_j$, that are about the $j$-th task. In the challenging setting we consider, the task identity is not known at test time, so that, after the training session $\mathcal{S}_t$, we transform a test sample $x$ into $\left\{\tilde{x}_j=g\left(x, \theta_{g_j}^{(j)}\right),\ j = 1,\ldots,t\right\}$.
In the {\it class-incremental} case, after the training session $\mathcal{S}_t$, we then compute $\left\{y_j = \hat{c}_j\left(m\left(\tilde{x}_j, \theta_m^{(t)}\right), \theta_{\hat{c}_j}^{(j)}\right),\ j = 1,\ldots,t\right\}$, and we concatenate the $y_j$'s to get a vector of confidence scores involving all the classes explored so far, that is used to make the final prediction on $x$ (recall that each $y_j$ is only about a subset of classes). 
Differently, in the {\it domain-incremental} case, after the training session $\mathcal{S}_t$ we compute $\left\{y_j = \hat{c}\left(m\left(\tilde{x}_j, \theta_m^{(t)}\right), \theta_{\hat{c}}^{(t)}\right),\ j = 1,\ldots,t\right\}$, and for each class we take the maximum confidence score to get the final decision vector (recall that here each $y_j$ is about all the classes, since the set of classes is shared by all the tasks). 
Since all these operations can be run in parallel over the different transformations (up to the final $y_j$'s), we refer to such classification procedure with the term \emph{parallel classifier} (see Fig. \ref{fig:2b}). 

\section{Experiments}\label{sec:experiments}
We describe our experimental investigation by introducing the considered datasets in Section~\ref{sec:data}, competitors in Section~\ref{sec:competitors}, neural architectures and experimental setup in Section~\ref{sec:setup}, and by reporting and discussing  results in Section~\ref{sec:results}, followed by an in-depth analysis of proposed approach.
\subsection{Datasets}\label{sec:data}
In order to assess the performance of the proposed learning algorithm, we exploit multiple datasets available in the literature. 
\cifards \cite{krizhevsky_learning_2009} is a popular Image Classification dataset, consisting of $60$k $32 \times 32$ color images from 100 different classes (500 training images per class). RESISC45 dataset \cite{resisc} is a benchmark for Remote Sensing Image Scene Classification (RESISC). This dataset contains $32$k color images, covering 45 scene classes (560 training images for class).
\fiveds \cite{ebrahimi2020adversarial} is the concatenation of five well-known image
classification datasets: CIFAR-10 \cite{krizhevsky_learning_2009}, MNIST\cite{lecun1998mnist}, Fashion-MNIST \cite{xiao2017fashion}, SVHN \cite{netzer2011reading}, and notMNIST\cite{notmnist}. Even though each single benchmark is individually fairly easy when using pretrained models, \fiveds is challenging because of the forgetting arising from very different distributions in the input space.
DomainNet \cite{peng2019moment} is a collection of images labelled over 345 classes with multiple drastic domain shifts. In this paper, we exploit 4 subsets (sketch, real, painting, clipart), totalling 250k training samples and 110k test samples.

We define four different CL problems using the just described four datasets: ({\it i.}) \cifartask, where CIFAR100  is randomly divided into 10 tasks (each task contains 10 classes) ({\it ii.}) \resisctask with 9 tasks (each task contains 5 classes); ({\it iii.}) \fivedstask, where each task consists of the 10 classes available in each of the subdatasets populating \fiveds, and  ({\it iv.}) \domnettask, where each task contains new examples of the same set of classes but from a different domain. While ({\it i.}, {\it ii.}, {\it iii.}) are {\it class-incremental}, ({\it iv.}) is {\it domain-incremental}. Moreover, ({\it i.}, {\it ii.}) are \singlesource, while  ({\it iii.}, {\it iv.}) are \multisource.

\subsection{Competitors}\label{sec:competitors} 
We compare the proposed \itpad and \itadd with a baseline model that trains only the classification head (parameters $\theta_{\hat{c}}$), \bt, and \ft{} (two cases, described below). In \itpad, \itadd and the baseline model, the learning process is driven by a loss that involves the supervised data available in each session, i.e., $\mathcal{D}_j$. Differently, in the case of \ft, the loss is augmented with CL regularizers from state-of-the art approaches.\footnote{As we will show in Section \ref{sec:indepth}, fine-tuning without any CL regularizers is impractical due to strong forgetting.} Our goal is to investigate whether IT is a competitive strategy without changing the loss function and without introducing extra computations.
As a matter of fact, a variety of different specific CL techniques have been developed in the last decade \cite{delange2021clsurvey}. 
In particular, data regularization methods are mostly inspired by knowledge distillation \cite{hinton2014distilling}. Learning without Forgetting (LwF) \cite{lwf} is the simplest instance and it basically performs distillation on output logits computed on 
 the currently available data, from the model obtained at the end of session $\mathcal{S}_{t-1}$ to the current model in $\mathcal{S}_t$. 
Clearly, distillation is restricted to the units of the classes learnt up to time $\mathcal{S}_t$. 
Alternatively, Learning without Memorizing (LwM) \cite{lwm} is geared towards attention. Specifically, it consists of an additional loss used to preserve attention maps 
over different sessions (see \cite{masana} for details). 
Finally, we also consider regularization approaches that directly target the weights, such as Elastic Weight Consolidation (EWC) \cite{ewc}. The importance of the weights is estimated through an approximation of the Fisher Information Matrix, and EWC enforces regularity on the learnable parameters between $\theta_{*}^{(t-1)}$ and $\theta_{*}^{(t)}$ according to such estimated importance. Authors of \cite{pathint} proposed an even simpler instance of EWC, since they show that the importance of each parameter to the fulfilment of a learning task can be estimated by accumulating individual weight changes; such method is called Path Integral in the following (also known as Synaptic Intelligence).

\subsection{Experimental Setup}\label{sec:setup}  
We focus on a specific class of convolutional networks for image classification, the widely popular ResNets \cite{resnet}, following our edge-oriented scenario in which the computational budget is indeed limited. In particular, we will consider the rather small ResNet-18 (11M parameters), pretrained on ImageNet \cite{imagenet} (input at the resolution of $224 \times 224$).
Differently from a Transformer-based architecture, it requires smaller memory and it is less computationally demanding (compared to those instances of Transformer models with a relatively smaller number of parameters - such as ViT-B/16, 84M parameters). 
In the following we consider two different partial finetuning settings, \ft1 and \ft2, focusing on the last module of the considered architecture, which is made of the two last BasicBlocks\footnote{Please refer to the PyTorch implementation of the ResNet architecture,  \url{https://pytorch.org/vision/0.8/_modules/torchvision/models/resnet.html}
for further details.}  and represents a remarkable fraction (70\% of the total parameters) of the whole architecture.
In \ft1 we restrict the tuning operations to the parameters  contained in the last BasicBlock (4.7M), and we refer to these parameters with $\theta'_{m}$. In \ft2 we also include the ones in the penultimate BasicBlock (+3.7M), indicating with $\theta''_{m}$ the union of these parameters. 

In general, we use cross-entropy as classification loss function, and the Adam optimizer is exploited for all the learnable parameters unless otherwise specified; a smaller batch size $B$ is employed for the smaller datasets ($B=16$ for CIFAR100, RESISC45; $B=64$ for FIVEDS, DOMAINNET). In all the experiments, networks are randomly initialized, using the same seed for the different approaches, and we report results averaged over 3 runs with different initializations. Following \cite{buzzega}, training for multiple epochs decouples the effects of forgetting and underfitting. As such, unless otherwise stated, we select a sufficient number of epochs to obtain a stable configuration of the parameters at the end of each task.
Since we assume to have a limited computational budget, hyper-parameters are shared across the different learning problems (i.e., not tuned specifically for the dataset at hand). In all the \itpad experiments we learn a 32-pixel thick border. Concerning CL strategies \footnote{Refer to the original papers for details on the role of these parameters.}, we adopt parameters suggested by the respective authors and further suggested by \cite{masana}: for LwF we set the temperature to $2$; for EWC the fusion of the old and new importance weights is done with $\alpha = 0.5$; for LwM we set $\beta=\gamma = 1.0$; for Path Integral we fix the damping parameter to 0.1 as proposed in the original work. 
Concerning batch normalization layers, we employ running averages at test time for \singlesource and fixed pretraining statistics for \multisource. While the impact on the final metrics is modest, it can be easily grasped that, in the case of very heterogeneous data across different tasks, learning the tuning of the model with fixed statistics results in less task-specific adaptation, hence less forgetting. On the other hand, if all the data are relatively homogeneous in terms of global statistics (e.g. CIFAR100), adapting the model in running mode could benefit from slightly more representative statistics with respect to the pretraining domain.

We measure the average accuracy and average forgetting at the end of the learning sequence (\ie \xspace $t=T$). The average accuracy $\bar{a}_{T}$ at the end of the considered task sequence is selected as the main metric, as in most of the continual learning literature. Let be $\tau$ the task/session index, $a_{t, \tau}$ is the accuracy computed on the test set of task $\tau$, with the model obtained after training on task $t$ (i.e., with parameters $\theta_{*}^{(t)}$). Average forgetting $\bar{f}_{T}$ is also helpful to evaluate the average magnitude of accuracy drops over the sequence.  Formally,
\begin{equation*}
\begin{aligned}
&\bar{a}_{t}=\frac{1}{t} \sum_{\tau=1}^{t} a_{t, \tau}, 
&\bar{f}_{t}=\frac{1}{t-1} \sum_{\tau=1}^{t-1} \max _{\tau^{\prime} \in\{1, \cdots, t-1\}}\left(a_{\tau^{\prime}, \tau}-a_{t, \tau}\right).
\end{aligned}
\end{equation*}
In all the following results, the terms {\it accuracy} and {\it forgetting} refer to the aforementioned average values.

\subsection{Results and Discussion}\label{sec:results}
The results of our experimental activities are reported in Tab.~\ref{tab:allres} and discussed in the following.

In the case of {\cifartask (\singlesource)}\xspace the domain shift from one task to another one in the sequence is relatively small and qualitatively all the data share similar visual features \cite{PARISI201954}. 
The first noteworthy remark is that the baseline, which only learns the classifier $\hat{c}$ with the label trick, works surprisingly well with respect to 
 fine-tuning paired with well-known CL regularizers. Interestingly enough, bias tuning (\bt) shows no practical improvement over the baseline, given the forgetting due to the incremental nature of the learning problem. The same trend can be observed for the partial fine-tuning options (\ft1, \ft2). On the contrary, both the IT approaches have at least the same accuracy as the baseline, showing that they are an appropriate way to strive for better performance, with slight additional complexity on top of the label trick baseline. While the improvement provided by \itadd is marginal, \itpad significantly improves the test accuracy without being particularly exposed to forgetting compared to the other options. As expected (Sec. \ref{sec:proposal}), the \emph{parallel classifier} approach does not help due to relative data homogeneity of different tasks (\singlesource).

\begin{table*}[h]
\begin{center}
\hskip -1mm\scalebox{0.94}{
\begin{tabular}{@{\hskip -0.005cm}l|l|c@{\hskip 0.1cm}c|c@{\hskip 0.1cm}c|c@{\hskip 0.1cm}c|c@{\hskip 0.1cm}c@{\hskip -0.005cm}}
\toprule
\textbf{Tuning} & \textbf{Learnt Parameters} & \multicolumn{2}{c|}{\textbf{\cifartask}} & \multicolumn{2}{c|}{\textbf{\resisctask}} & \multicolumn{2}{c|}{\textbf{\fivedstask}}& \multicolumn{2}{c}{\textbf{\domnettask}}\\
 &  & {Accuracy} $\uparrow$ & {Forgetting} $\downarrow$ & {Accuracy} $\uparrow$ & {Forgetting} $\downarrow$ & {Accuracy} $\uparrow$ & {Forgetting} $\downarrow$ & {Accuracy} $\uparrow$ & {Forgetting} $\downarrow$ \\ \midrule
None & $\theta_{\hat{c}}$ & 44.49 {\tiny $\pm 0.22$} & 13.92 {\tiny $\pm 1.49$} & 57.14 {\tiny $\pm 0.50$} & 22.51 {\tiny $\pm 1.15$} & 44.20 {\tiny $\pm 2.79$} & 17.78 {\tiny $\pm 1.30$}& 35.93 {\tiny $\pm 0.13$} & 18.98 {\tiny $\pm 0.32$}\\
\midrule
\bt & $\theta_{\hat{c}}$ and Biases of $m$ ($\sim$ 1k) & 41.72 {\tiny $\pm 0.30$} & 29.91 {\tiny $\pm 0.81$} & 46.79 {\tiny $\pm 2.06$} & 35.24 {\tiny $\pm 1.19$}& 20.36 {\tiny $\pm 2.86$} & 56.92 {\tiny $\pm 2.11$}& 38.85 {\tiny $\pm 1.93$} & 23.28 {\tiny $\pm 2.31$}\\
\midrule
\ft1-LwF& $\theta_{\hat{c}}$ and $\theta'_{m}$ (4.7M) & 39.18 {\tiny $\pm 0.83$} & 24.62 {\tiny $\pm 0.65$} & 54.46 {\tiny $\pm 1.15$} & 12.07 {\tiny $\pm 1.41$}& 52.72 {\tiny $\pm 1.15$} & 28.85 {\tiny $\pm 1.70$} & -- & -- \\
\ft1-LwM& $\theta_{\hat{c}}$ and $\theta'_{m}$ (4.7M) & 38.26 {\tiny $\pm 0.72$} & 22.19 {\tiny $\pm 0.81$} & 52.51 {\tiny $\pm 1.06$} & 25.92 {\tiny $\pm 1.49$}& 29.34 {\tiny $\pm 0.85$} & 67.02 {\tiny $\pm 4.05$} & -- & -- \\
\ft1-EWC& $\theta_{\hat{c}}$ and $\theta'_{m}$ (4.7M)  & 39.62 {\tiny $\pm 0.48$}& 22.85 {\tiny $\pm 0.73$} & 54.17 {\tiny $\pm 0.45$}& 25.75 {\tiny $\pm 1.46$}& 38.45 {\tiny $\pm 2.02$} & 55.18 {\tiny $\pm 0.76$} & 19.05 {\tiny $\pm 1.77$} & 9.72 {\tiny $\pm 0.71$}\\
\ft1-PathInt& $\theta_{\hat{c}}$ and $\theta'_{m}$ (4.7M) & 37.46 {\tiny $\pm 1.22$}& 23.49 {\tiny $\pm 0.97$} & 52.95 {\tiny $\pm 1.64$} & 24.47 {\tiny $\pm 1.35$}&51.38 {\tiny $\pm 1.34$} & 25.12 {\tiny $\pm 0.52$} & 38.55 {\tiny $\pm 1.48$} & 5.7 {\tiny $\pm 1.86$}\\
\midrule

\ft2-LwF& $\theta_{\hat{c}}$ and $\theta''_{m}$ (8.4M) & 43.03 {\tiny $\pm 0.49$} & 28.61 {\tiny $\pm 0.22$} & 53.78 {\tiny $\pm 0.45$} & 29.27 {\tiny $\pm 2.05$}& $\textbf{61.45}$ {\tiny $\pm 3.02$} & 28.68 {\tiny $\pm 0.23$} & -- & --\\
\ft2-LwM& $\theta_{\hat{c}}$ and $\theta''_{m}$ (8.4M) & 42.06 {\tiny $\pm 0.63$} & 28.57 {\tiny $\pm 1.04$} & 53.94 {\tiny $\pm 0.88$} & 28.94 {\tiny $\pm 2.60$}& 40.98 {\tiny $\pm 0.87$} & 35.71 {\tiny $\pm 2.45$} & -- & --\\
\ft2-EWC& $\theta_{\hat{c}}$ and $\theta''_{m}$ (8.4M) & 41.91 {\tiny $\pm 0.81$} & 26.76 {\tiny $\pm 0.64$} & 56.03 {\tiny $\pm 1.39$} & 27.17 {\tiny $\pm 0.41$}& 46.25 {\tiny $\pm 0.88$} & 53.11 {\tiny $\pm 1.05$}& 29.59 {\tiny $\pm 1.31$} & 11.21 {\tiny $\pm 1.42$}\\
\ft2-PathInt& $\theta_{\hat{c}}$ and $\theta''_{m}$ (8.4M) & 42.11 {\tiny $\pm 1.08$} & 27.27 {\tiny $\pm 1.01$} & 54.98 {\tiny $\pm 1.86$} & 26.02 {\tiny $\pm 0.80$}& 60.24 {\tiny $\pm 2.05$} & 29.09 {\tiny $\pm 2.62$}& $\textbf{39.41}$ {\tiny $\pm 1.65$} & 16.17 {\tiny $\pm 3.68$}\\
\midrule
\midrule
\itpad (Standard) & $\theta_{\hat{c}}$ and {\it Frame} (0.1M) & $\textbf{51.58}$ {\tiny $\pm 1.78$} & 19.31 {\tiny $\pm 1.46$} & $\textbf{61.06}$ {\tiny $\pm 0.99$} & 16.99 {\tiny $\pm 0.94$}& 44.65 {\tiny $\pm 1.23$} & 31.86 {\tiny $\pm 0.89$}& 38.74 {\tiny $\pm 0.44$} & 18.85 {\tiny $\pm 1.39$}\\
\itadd (Standard) & $\theta_{\hat{c}}$ and {\it Perturbation} (0.15M) & 44.81 {\tiny $\pm 0.83$} & 21.01 {\tiny $\pm 0.85$} & 57.18 {\tiny $\pm 1.63$} & 19.43 {\tiny $\pm 0.64$} & 36.09 {\tiny $\pm 1.85$} & 44.29 {\tiny $\pm 0.97$}& 33.67 {\tiny $\pm 0.30$} & 22.45 {\tiny $\pm 0.34$}\\
\midrule
\itpad (Parallel) & $\theta_{\hat{c}}$ and {\it Frame} (0.1M/task) & 46.54 {\tiny $\pm 0.82$} & 12.88 {\tiny $\pm 2.27$} & 55.84 {\tiny $\pm 1.59$} & 19.68 {\tiny $\pm 1.81$} & 53.36 {\tiny $\pm 0.55$} & 19.97 {\tiny $\pm 1.40$}& $\textbf{43.93}$ {\tiny $\pm 1.22$} & 16.12 {\tiny $\pm 0.91$}\\
\itadd (Parallel) & $\theta_{\hat{c}}$ and {\it Perturb.} (0.15M/task) & 41.82 {\tiny $\pm 1.23$} & 13.05 {\tiny $\pm 1.14$} & 52.47 {\tiny $\pm 2.83$} & 18.29 {\tiny $\pm 2.51$} & $\textbf{56.32}$ {\tiny $\pm 0.43$} & 17.75 {\tiny $\pm 0.37$}& 41.18 {\tiny $\pm 0.83$} & 17.34 {\tiny $\pm 1.61$} \\
\bottomrule
\end{tabular}}
\caption{Average accuracy ($\uparrow$, higher is better) and forgetting ($\downarrow$, {lower is better}) measured at the end of the learning sequence.  
Notice that some CL strategies are not well-suited for the domain-incremental setting (see the paper text; invalid configurations are then marked with ``--'').  
In the second column we report the set of parameters subject to optimization: we always learn the classification head; we tune network weights in the \bt, \ft \xspace approaches and we learn transformation parameters (\emph{Perturbation} or \emph{Frame}) in the IT approaches. We report the number of such learnt parameters in brackets.}\label{tab:allres}
\end{center}
\end{table*}

In {\resisctask (\singlesource)}, one may wonder whether a relatively high semantic affinity between the pre-training domain (ImageNet) and the target tasks is crucial in order to get improvements with respect to the baseline.  
In fact, RESISC45 is a publicly available dataset of satellite imagery, which features a quite large semantic and perceptual shift with respect to the pre-training domain.  
Similarly to the previous case, the regularization-based CL strategies are struggling in achieving and maintaining a good performance throughout the task sequence, dropping to accuracy levels that are lower than the baseline. On the other hand, \itpad is again the best performing option and provides an appreciable accuracy increase, while being less affected by forgetting than the other tuning options (excluding \ft1-LwF, characterized by a consistently lower accuracy though).

In {\fivedstask (\multisource)}  we experiment with data coming from multiple sources, possibly distant in the semantic and perceptual spaces.  
Interestingly enough, in this case LwF provides the highest performance. Given that it has been originally proposed as a task-incremental strategy it is not surprising that it performs well on a sequence of very distinct tasks (both in the semantic and in the perceptual spaces). At the same time, we can see that \itpad and \itadd provide valuable improvement when implemented in the {\it parallel classifier}. Moreover, compared to LwF and Path Integral, the amount of learnt parameters (and the computational burden) is smaller, there is no need to store tensors of the same size as the weights (importance weights for PathInt, model snapshot for LwF) and forgetting is lower. 

The case of {\domnettask (\multisource)} departs from the previous ones, being it a domain-incremental setting. The semantic space is shared by all the tasks, which feature remarkably different visual styles and perceptual features (color, texture, etc.). Although the obtained accuracy may seem a bit low, it is important to remark that it is a very challenging learning problem, with many examples that are hard even for humans and especially for convolutional networks, that are known to heavily rely on texture \cite{geirhos2018imagenettrained}. We did not apply LwF and LwM, that do not fit well the domain-incremental setting. In fact, the knowledge distillation term would call for the network output with new data to (a) fit the classification loss and (b) be similar to the output of the model learnt at the previous task, which is not expected to help in effectively learning with low forgetting. 
Also, it should be noted that in this case the baseline itself cannot exploit the label trick, given that each task contains data for the entire class set. As such, several methods offer improvements over the baseline, including the quite simple bias tuning (\bt). Moreover, Path Integral gives also a similar improvement, although implying to compute and store parameter-specific importance weights. On the other hand, it can be clearly appreciated that the proposed \itpad implemented with {\it parallel classifier} features the highest accuracy, confirming the importance of learning independent transformations in \multisource. 

\section{In Depth Analysis}\label{sec:indepth}
We perform additional experiments aimed at gaining more insights on the previously described results, focusing on the \cifartask learning problem.

\begin{figure}[h]
\includegraphics[width=0.35\textwidth]{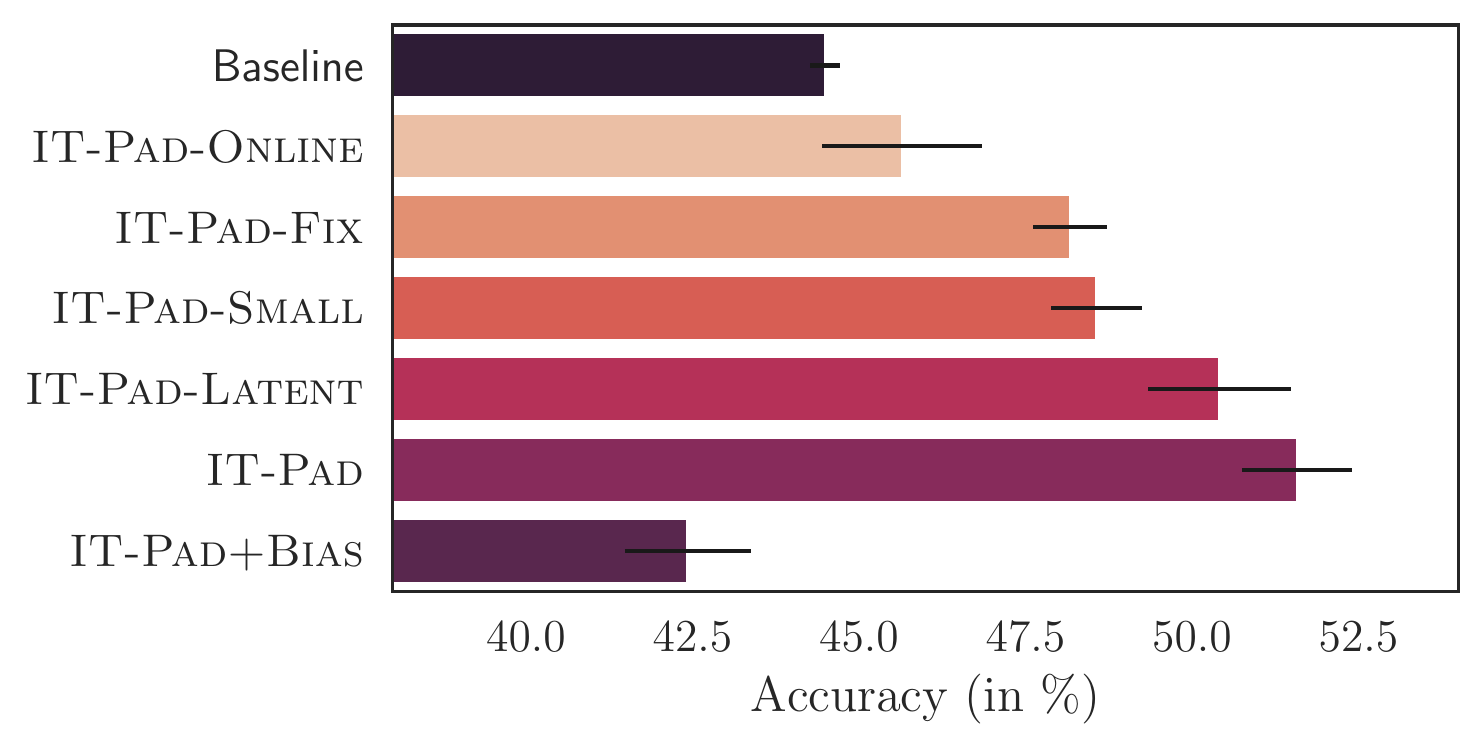}
\centering
\caption{Average test accuracy measured at the end of the learning sequence, for variants of \itpad in \cifartask. Performing the transformation in some latent space (instead of the input) is under-performing and adding bias tuning seems to be counterproductive.
}
\label{fig:cifarablation}
\end{figure}

In Fig. \ref{fig:cifarablation} we report the accuracy obtained in a comparative experiment with 5 different variants of the \itpad scheme. ($i.$) \itpad-Online is the setting in which training is performed with a single pass on the training data. This speeds up learning but it greatly reduces the extent of the improvement at the end of training (compared to Tab.~\ref{tab:allres}). ($ii.$) \itpad-Fix refers to the setting in which we kept fixed the \textit{Frame} (the learnable padding) after the first task. Results show that learning the additional pixels is really beneficial only if they are allowed to adapt to the slight variations of the different tasks. ($iii.$) \itpad-Small is obtained reducing by a factor of 4 the thickness of the frame border, decreasing the total amount of additional parameters by a factor of 5; it is interesting to notice that the accuracy is 4\% higher than the Baseline (first row of Tab.~\ref{tab:allres}) also in that setting. ($iv.$) \itpad-Latent is about applying the padding operation in a latent space (right after the first two convolutional layers) and benefits of a comparable improvement with a similar amount of additional learnable parameters ($<0.1$M). On the other hand, ($v.$) combining the most promising IT approach with \bt (\itpad+Bias), completely vanishes any improvement, coherently with what was suggested in \cite{vpt}.
This analysis confirms the value of the simple-but-effective vanilla \itpad scheme.

In Tab. \ref{tab:cifaronline}, we further investigate the online setting, showing the test accuracy for the different methods. In general, the gap between the \itpad approach and all the competitors is even wider with respect to the multi-epoch setting, given that fine-tuning a larger portion of the network would require a larger amount of update steps in order to obtain a stable configuration of the weights, conflicting with the single-epoch constraint. 
\begin{table}[h]
\begin{center}
\scalebox{0.94}{
\begin{tabular}{@{\hskip -0.01cm}l|l|c|c@{\hskip -0.01cm}}
\toprule
\textbf{Tuning} & \textbf{Learnt Parameters} & Accuracy $\uparrow$ & Forgetting $\downarrow$ \\ \midrule
None & $\theta_{\hat{c}}$ & 41.38 {\tiny $\pm 0.56$} & 14.11 {\tiny $\pm 1.88$}\\
\midrule
\bt & $\theta_{\hat{c}}$ and Biases ($\sim$1k) & 43.94 {\tiny $\pm 2.31$} & 24.87 {\tiny $\pm 3.18$}\\
\midrule
\ft1 (LwF)& $\theta_{\hat{c}}$ and $\theta'_{m}$ (4.7M) & 26.07 {\tiny $\pm 0.41$} & 20.19 {\tiny $\pm 0.92$}\\
\ft1 (LwM)& $\theta_{\hat{c}}$ and $\theta'_{m}$ (4.7M) & 26.09 {\tiny $\pm 0.64$} & 20.08 {\tiny $\pm 0.68$}\\
\ft1 (EWC)& $\theta_{\hat{c}}$ and $\theta'_{m}$ (4.7M)  & 25.88 {\tiny $\pm 0.83$}& 23.11 {\tiny $\pm 0.77$}\\
\ft1 (PathInt)& $\theta_{\hat{c}}$ and $\theta'_{m}$ (4.7M) & 26.06 {\tiny $\pm 1.98$}& 20.84 {\tiny $\pm 1.41$}\\
\midrule

\ft2 (LwF)& $\theta_{\hat{c}}$ and $\theta''_{m}$ (8.4M) & 29.13 {\tiny $\pm 0.71$} & 22.48 {\tiny $\pm 0.86$}\\
\ft2 (LwM)& $\theta_{\hat{c}}$ and $\theta''_{m}$ (8.4M) & 22.84 {\tiny $\pm 0.55$} & 31.83 {\tiny $\pm 1.29$}\\
\ft2 (EWC)& $\theta_{\hat{c}}$ and $\theta''_{m}$ (8.4M) & 23.14 {\tiny $\pm 0.32$} & 21.32 {\tiny $\pm 0.58$}\\
\ft2 (PathInt)& $\theta_{\hat{c}}$ and $\theta''_{m}$ (8.4M) & 26.69 {\tiny $\pm 1.72$} & 24.85 {\tiny $\pm 1.95$}\\
\midrule
\midrule
\itpad (Standard) & $\theta_{\hat{c}}$ and {\it Frame} (0.1M) & $\textbf{45.65}$ {\tiny $\pm 1.20$} & 17.89 {\tiny $\pm 1.92$}\\
\itadd (Standard) & $\theta_{\hat{c}}$ and {\it Perturb.} (0.15M) & 42.48 {\tiny $\pm 1.34$} & 14.87 {\tiny $\pm 0.82$} \\
\bottomrule
\end{tabular}}
\caption{Results on \cifartask in the online setting: average accuracy and forgetting are measured at the end of the learning sequence. }\label{tab:cifaronline}
\end{center}
\end{table}

\begin{figure}[h]
\includegraphics[height=5.5cm,trim={3.0mm 0 1mm 0},clip]{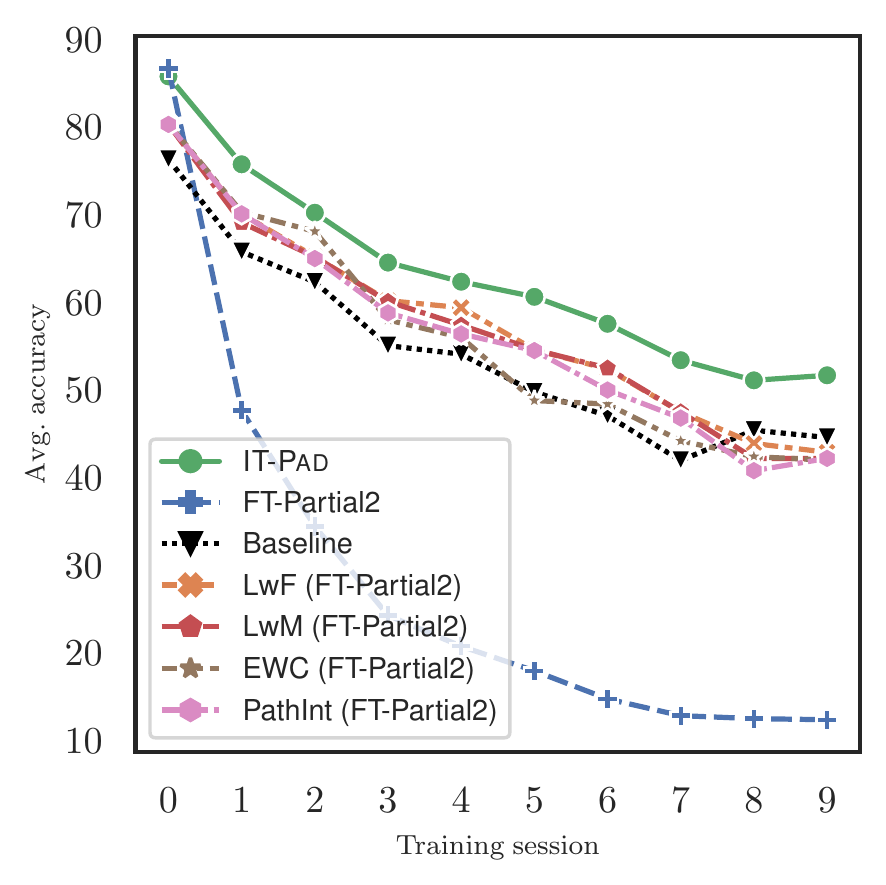}\includegraphics[height=5.45cm,trim={1mm 0 1.5mm 0},clip]{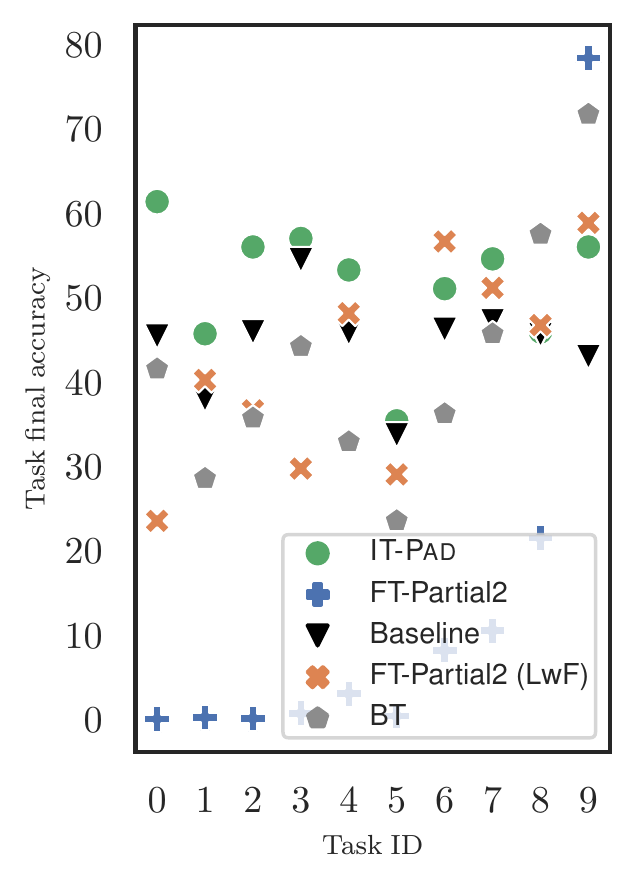}
\centering
\caption{\cifartask, models from Tab. \ref{tab:allres}. Left: Average accuracy during the entire learning sequence. The blue curve (\ft2, with no CL regularizers) highlights the fact that naively fine-tuning is not a practical option. \itpad (green) shows the best behavior throughout all the learning sequence. Right: Task-specific accuracy, at the end of the learning sequence. Fine-tuning without CL regularizers (\ft2, blue) has extremely low accuracy, excluding the very last task. \bt (grey) and \ft2 (LwF, orange) best-performing tasks are concentrated in the last part of the  sequence (task id $\geq 6$). \itpad consistently beats the baseline and is the best for most of the tasks.}
\label{fig:cifardepth}
\end{figure}

In Fig. \ref{fig:cifardepth} we provide some insights on the test accuracy, measured during the learning sequence and at the end of it, respectively.
In Fig. \ref{fig:cifardepth} (left), we show that the \itpad approach typically has the highest average accuracy throughout the learning sequence, while using \ft{} without any further CL regularizers is not a viable option. In Fig. \ref{fig:cifardepth} (right) we can see that the task accuracy of \itpad is pretty uniform over the entire set (with no drastic forgetting on the old tasks) and generally the highest among the considered approaches.

\section{Conclusions}\label{sec:conclusions}
We presented Input Tuning, a novel tuning procedure for the exploitation of pre-trained models in the context of continual learning by tweaking the input data (especially when inserting a frame of learnable pixels, referred to as \itpad). We empirically showed that the proposed method is simple but effective in consistently improving the quality of the outcome on multiple learning problems, and can be promptly extended to the \multisource case. While the visible improvements can be intuitively connected with findings about the prevalence of forgetting in the last layers of the network \cite{lastlayer, ramasesh2021anatomy}, we plan to get deeper insights into the learning dynamics of the proposed method.

\bibliography{biblio}
\bibliographystyle{IEEEtran.bst}


\end{document}